# Comparative Analysis: Violence Recognition from Videos using Transfer Learning

Dursun Dashdamirov
School of IT and Engineering
ADA University, Baku, Azerbaijan
Department of Computer Science
George Washington University, Washington, D.C., the U.S.
0009-0005-3590-8857

*Abstract*—*Action recognition has become a hot topic in computer vision. However, the main applications of computer vision in video processing have focused on detection of relatively simple actions while complex events such as violence detection have been comparatively less investigated. This study focuses on the benchmarking of various deep learning techniques on a complex dataset. Next, a larger dataset is utilized to test the uplift from increasing volume of data. The dataset size increase from 500 to 1,600 videos resulted in a notable average accuracy improvement of 6% across four models.*

*Keywords—video classification, violence detection, transfer learning, benchmarking*

## I. Introduction

One of the most threatening violations of normal social habits is violence. Violence has been observed throughout human history. It can be considered as part of daily life which includes incidents of aggressive behavior towards children in educational settings, such as schools and kindergartens, as well as instances of domestic violence within households, road rage incidents, and public altercations. The police and security assigned to the designated area endeavor to prevent and/or punish participants of this act. However, due to the shortage and inattention of personnel, these actions may not be prevented in a timely manner. Even when it does, this process requires huge amounts of personnel to be designated for surveillance of the area.

The widespread deployment of surveillance cameras enabled authorities to detect and respond to violent incidents swiftly, thereby mitigating their catastrophic repercussions. However, the majority of contemporary surveillance systems necessitate manual human scrutiny of video footage to identify such scenarios—a task that is both impractical and inefficient. Therefore, Computer Vision has the potential to automate this process. The CCTV footage can be fed into a robust algorithm that can detect violence between human beings in real-time. This will eliminate the cost of surveillance and will mitigate privacy concerns since there will not be human involvement in the process.

However, considerable effort must be expended to guarantee the effectiveness and efficiency of these algorithms. Numerous challenges must be addressed to enhance the applicability of the computer vision algorithms. Firstly, the footages vary significantly in their context. A large amount of training data must be collected and labeled to train effective algorithms. Today, there are available public video datasets. Nonetheless, an expansion in the volume of these datasets remains critically essential.

Secondly, the resolution of the videos, frame rate, lighting conditions, and perspective of the camera vary significantly. These differences make the development of robust and generalized models more challenging. Additionally, the length of the violent act and its reflection on the footage are also important. The videos may contain occlusions and obstructions. These obstructions, combined with dynamic backgrounds can hinder accurate detection.

Another bottleneck in the solution process of the problem is computational need. Transfer learning algorithms leverage the knowledge collected from solving one problem to solve a different but related problem, by reusing pre-trained models as the initial point on a new task. This approach can significantly reduce the time and data required for model development, enabling efficient adaptation to new tasks with reasonably minimal effort.

Considering the challenges and the inherent complexity of the problem, this research aims to address two main questions: Can pre-trained models achieve more accurate results compared to traditional deep learning algorithms, and to what extent does more data improve accuracy?

To achieve this, a comparative analysis is conducted to identify the most robust algorithm. Then, this algorithm is trained on a larger dataset to understand the uplift from the usage of larger datasets for complex computer vision tasks.

The remainder of the paper is organized as follows. Section 2 gives background about the problem we are trying to solve and gives an overview of the related work. In section 3, the technical details of the proposed analysis are explained. This section also includes details about the pre-processing of the data. The paper ends with conclusions and recommendations for future work. All utilized in-text references are listed at the end of the paper.

## II. Related Work

There have been several studies attempting to detect violence from video. In the following sections, they are classified into two groups: Handcrafted feature-based and Deep learning-based approaches.

### A. Handcrafted feature-based approaches

Datta et al. [1] used motion trajectory and limb orientation data to identify violence. Nguyen et al. [2] proposed a hierarchical hidden Markov model (HHMM) which could be effective in recognizing aggressive behaviors, mainly applying a standard HHMM framework for violence detection. Kim and Grauman [3] employed a combination of probabilistic Principal Component Analysis (PCA) to capture local flow patterns and a Markov Random Field (MRF) to preserve global consistency. In contrast, Mahadevan et al. [4] claimed that optical flow-based representations are insufficient for detecting abnormalities in appearance and motion. They developed a technique for identifying violent scenes by monitoring various features such as blood, flames, motion intensity, and loudness.

---
[1] https://github.com/DDursun/Violence-Detection

## B. Deep learning-based approaches

Violent acts vary significantly in their nature. Handcrafted feature-based approaches are not the fastest and most accurate way of determining violent acts. There have also been considerable amount of work in the context of deep learning methods. The first remarkable approach in the deep learning domain was the usage of audio and visual data to solve the problem. Several studies [5,6,7] have been conducted with this combinatory approach. For audio, Lin et al. [7] trained a weakly supervised classification algorithm and joined it with a visual classification algorithm which is able to identify movement, explosions, and blood.

Serrano et al. [8] proposed the creation of a representative image by emphasizing major areas containing motion for the classification of fight and non-fight videos. This summarization technique is combined with a 2D Convolutional Neural Network (CNN) and utilized to process images for classification. This approach uses a combination of spatial features and also temporal information. Researchers achieved promising results by combining summarization and the strength of CNNs in handling complex patterns.

In recent works, more approaches containing LSTM network have been experimented. The LSTM demonstrated robust performance due to its ability to deal with vanishing gradient problem. Abdali et al. [9] combined CNNs for spatial feature extraction and LSTMs for learning temporal relationships. The model achieved an accuracy of 98% and a processing speed of 131 frames per second, outperforming earlier models in terms of both accuracy and speed across diverse video resources. LSTM also proved itself to be powerful when utilized in tandem with good encoders and feature extractors. Hanson et al. [10] attempted to solve this problem using Bidirectional Convolutional LSTM (BiConvLSTM) architecture. The BiConvLSTM architecture integrates LSTM networks capable of processing data in both directions: forward and reverse temporal. They achieved high accuracy in well-known datasets with the power of temporal encodings and power of the Bi-LSTM network.

Humans do not learn every action from scratch. We acquire knowledge on one task and use it on another. Considering that feature extraction is a costly operation in tasks such as video analytics, pre-trained models are also utilized in this domain [11,12,13]. Several studies have used transfer learning to classify violent and nonviolent video footage. Transfer learning is a machine learning method in which a model developed to solve one problem is utilized in the next problems. It involves leveraging pre-learned patterns to improve feature extraction and learning efficiency. Mumtaz et al. [12] used this approach with GoogleNet, a deep CNN model pre-trained on the ImageNet dataset which is later adapted for violence detection. By further fine-tuning on Hockey and Movies datasets, they achieved ~99% accuracy. The paper by Sernani et al. [14] tested transfer learning on an AIRTLab dataset [15]. Pre-trained 3D CNN is utilized to extract features from the videos and these features are fed into a Support Vector Machine (SVM) for classification. This work also achieved a reduction of false positives.

The work by Gadelkarim et al. [16] also employed 3D CNN architecture. In this work, 3D CNN is combined with pre-trained Inception V3 and Gated Recurrent Units. The architecture achieved promising results both on binary classification and the multi-class (various kinds of violence classes) classification problems in a diverse dataset.

Jain and Vishwakarma [17] also utilized a pre-trained Inception Resnet V2 model to detect violence. Their approach included RGB video to Dynamic Image conversion (DIs). The model demonstrated high accuracy in 3 different datasets: Hockey Fight Dataset (93.3%), Real-life Violence Dataset (87%) and Movie Dataset (100%).

An interesting cross-species learning approach was utilized by Yujun Fu et al. [18] as a transfer learning method. The study utilized common features of motion during the fight process of species. An ensemble learning technique is utilized to combine learning from various animal fight data sources. The effectiveness of this approach is demonstrated through its ability to achieve competitive accuracy rates, with adaptation accuracies from animal fights and other human scenarios at 88.9% and 88.4%, respectively. Using the similarities, the model can learn from animal behaviors to detect human fights. This cross-species learning not only suggests a promising avenue for overcoming data limitations in this context but also features the potential of transfer learning.

## III. METHODOLOGY

The following section discusses the proposed approach and followed methodology. The section starts by defining the violent act and description of the dataset that is used. Next, it explores the data preparation steps, including modeling of data, and applied transformations to the image sequences. Subsequently, utilized deep learning models and the details of the training process are explained. This part also includes how pre-trained algorithms are fine-tuned to the dataset.

### A. Definition of a violent act

Violence is defined as "The intentional use of physical force or power, threatened or actual, against oneself, another person, or against a group or community, that either result in or has a high likelihood of resulting in injury, death, psychological harm, maldevelopment or deprivation" the World Health Organization [19].

### B. Dataset

The data utilized in the training and prediction are called "Real life Violence Situations Dataset" [20]. The total size of the dataset is ~2 GB. The dataset consists of 2000 videos with an even split between violent and nonviolent classes (1000 violent and 1000 nonviolent videos). The videos that are labeled as "violent" contain street fights, and fights in crowds in several environments and conditions. The videos labeled as nonviolent contain daily life situations including sports, eating, scenes from movies, and casual conversations between individuals. This dataset is considered to be more complex compared to other well-known datasets used to solve this problem. The videos have different characteristics in terms of camera angle, lighting conditions, resolution, and frame rate. In Fig. 1, two videos from the dataset are visualized. The difference in aforementioned characteristics can be easily seen through rows of the visual. Also, the video footages vary significantly in their content. These differences imitate realistic scenarios. Hence, forming a challenge in terms of well-generalized and accurate model building.

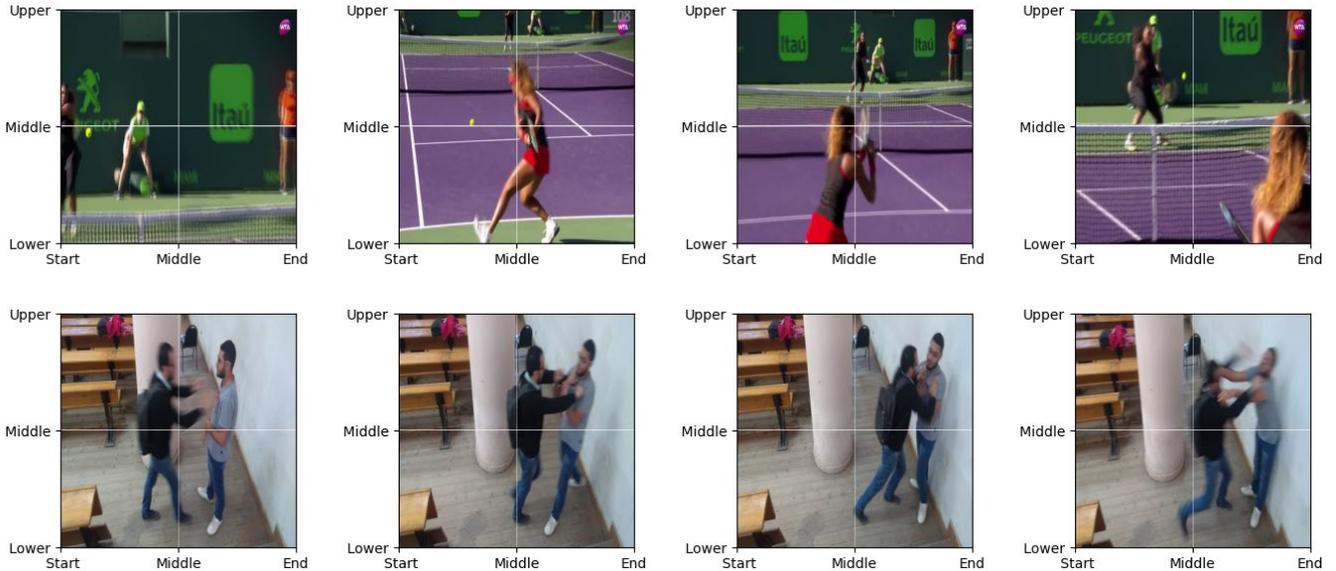

*Fig. 1. Four images (in row) from two videos of "Real-life violence situations' dataset. Evenly distributed four frames are extracted and visualized from original dataset. No augmentation technique is applied.*

## C. Data preparation

In almost all machine learning applications, the dataset needs to be preprocessed before it can be used by the model. This is particularly important considering the complexity of the format of the data – videos. In this process, raw images are transformed into arrays which enable the algorithms to process the data. The video is a sequence of images. It has to be converted into an array of numbers but before that step, we start by defining the dimensions of the images. This section explains how data are prepared for the model step-by-step.

All videos are resized into 100x100 in order to ensure consistency through the video shape and size. Also, this operation reduces the computational load and memory usage. The videos are segmented into sequences of 15 evenly spaced frames, selected to represent the entire 5-6 second duration. This approach ensures that key actions within the video are adequately captured, with approximately 3 frames per second.

Next, each extracted frame undergoes a series of image augmentation techniques. This process does not include the creation of new instances for training but augments the existing videos. This process prepares the model for the real-world challenges mentioned in Section I. The process starts by applying zoom to the images (randomly, 1-1.5 times). This simulates the object distance variability which is the case in real-world scenarios. The second technique involves adjusting the brightness of the image to ensure the model's robustness to varying lighting conditions (randomly, 0.8-1.5 times pixelwise multiplication). Since recorded videos are often unstable and low quality, Gaussian blurring (sigma value of 0.5-1.5) is applied to the images. This enhances the model's robustness with low-quality videos where the subject and movements are not clearly depicted.

After this step, we obtain the final dataset with 2000 videos. The experimentation process consists of two parts: using a fraction of data and using full dataset of 2000 videos. For both cases, 400 videos are separated from the dataset to ensure fair comparison of the effect of dataset size on accuracy. In the initial analysis with the smaller dataset, 500 training videos have been utilized while the figure is 1600 videos for the second part of the experiment. The final shapes of the datasets are tabulated in Table 1.

*Table 1. Shape of test and training datasets*

| Dataset | Number of Samples | Feature Dimensions | Labels Dimensions |
|---|---|---|---|
| Training-fraction | 500 | 15x100x100 | 2 |
| Training-full | 1600 | 15x100x100 | 2 |
| Test | 400 | 15x100x100 | 2 |

## D. Utilized models

Before discussing the results, this section will further discuss the four deep and transfer learning algorithms applied to solve this problem:

1. 3D CNN
2. 2D CNN and Bi-LSTM
3. InceptionV3 and Bi-LSTM
4. MobileNetV2 and Bi-LSTM

These four approaches have been identified for comparative analysis based on literature review. Further reasoning for selection is provided in each corresponding section.

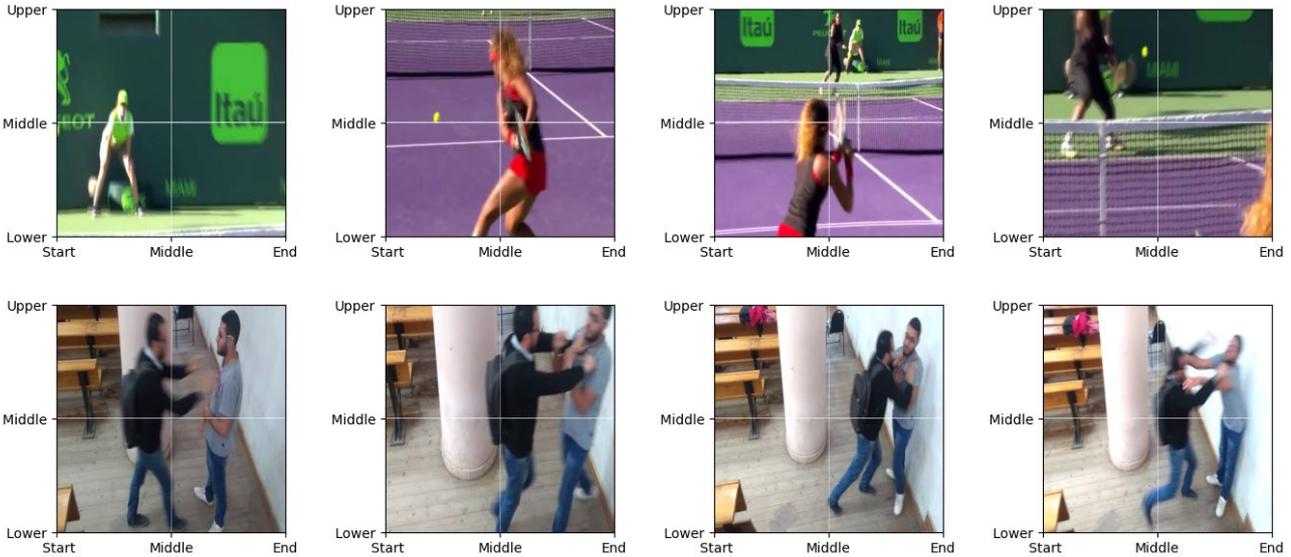

*Fig. 2. Four augmented images (in row) from two videos of "Real-life violence situations" dataset Evenly distributed four images are extracted, with all three augmentation techniques from section III.C applied to each frame within their respective ranges.*

*1) 3D CNN*

CNNs are a class of deep neural networks that are highly effective for processing data that have a grid-like topology (images etc.). CNNs employ layers with convolving filters that apply over local regions of the input. 3D architecture construct features from both spatial and temporal dimensions by performing 3D convolutions. This architecture is developed to work with volumetric data and image sequences. It is widely used in medical image analysis. The 3D CNNs are also found to be effective in human action recognition [21].

Our approach utilized two sets of 3D CNN layers. The first layer is a 3D convolution with 32 filters of size 3×3×3, using ReLU activation, with "same" padding, and L2 regularization. This is followed by a "MaxPooling3D" layer with a 2×2×2 pool size and Batch Normalization. Similar to the first a second 3D layer with 64 filters is added and it's followed by another "MaxPooling3D" and Batch Normalization layer. The model is then flattened, leading to a fully connected Dense layer with 256 units, ReLU activation, and L2 regularization, followed by a Dropout layer with a 0.5 rate. Finally, the output layer is a Dense layer with a "softmax" activation, corresponding to the number of classes for classification.

*2) 2D CNN and Bi-LSTM*

This approach is considered to be the baseline model for the next two approaches. The structure of the model is similar to the transfer learning methods and the only difference is the feature extraction methods. Here, CNNs are utilized to extract features from frames. Model uses two CNN layers with 64 and 128 filters, respectively, each followed by Batch Normalization and "MaxPooling2D". These layers are applied to each frame in the sequence. After flattening the 2D feature maps, Bidirectional LSTM layers capture temporal dependencies (more information provided in the next section). The "Dropout" layers are applied to prevent overfitting. The model concludes with Dense layers for classification, including two fully connected layers with "Dropout", and a final "softmax" layer for classification.

*3) Transfer Learning methods*

Long Short-Term Memory (LSTM) networks are a special kind of recurrent neural network (RNN) that are capable of understanding long-term dependencies in data. Bi-LSTMs improve this methodology by treating the data in both forward and backward directions. It is achieved by the usage of two separate hidden layers, which are then fed forward to the same output layer. The Bi-LSTM architecture is utilized in various domains including but not limited to video classification. Fig. 3. illustrates the simplified architecture of Bi-LSTM in the context of the problem.

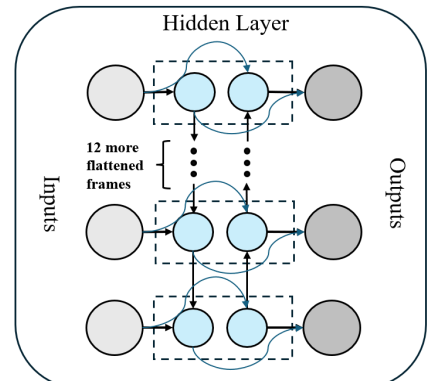

*Fig. 3. Architecture of Bi-LSTM network*

For last two approaches, transfer learning methods have been utilized in tandem with Bi-LSTM. These approaches included the usage of pre-trained models, namely, MobileNetV2 and InceptionV3.

InceptionV3 deep learning model architecture is developed by Google researchers. It is particularly suited for image recognition problems. This model is quite complex and computationally demanding compared to the MobileNetV2. InceptionV3 introduced the concept of "modules" containing parallel convolutional layers with various kernel sizes. This allows the model to capture spatial hierarchies at different scales.

The second transfer learning method utilized is combination of MobileNetV2 and Bi-LSTM. MobileNetV2 is an efficient CNN architecture designed primarily for mobile devices. MobileNetV2 introduced the concept of inverted residuals and linear bottlenecks. These features allow the network to leverage lightweight depth-wise convolutions to filter features in the intermediate expansion layer. This causes more complex features to be captured before narrowing down the dimensions of channels. This architecture finds an optimal balance between accuracy and computational efficiency making it a perfect fit for the cases of limited resources.

Both models are initialized with weights of the "ImageNet" dataset. This provides a good starting point for feature extraction. The models are set to be partially trainable. Several configurations have been tested and fine tuning the last 80 layers were found to work best for the given problem. This allowed algorithms to preserve pre-learned generic knowledge while adapting to the specific dataset..

*E. Training details*

To get a sense of best hyperparameters for the models, three main parameters are tuned on the "training-fraction" dataset. The range of parameters included choice of optimizer, learning rate and batch size. The validation dataset of 125 videos are utilized out of 500 videos. Then, these hyperparameters are used in both training processes.

During the training process, two main learning rate decay techniques are experimented. The transfer learning methods obtained better results with custom scheduler that used exponential decay with decay rate of 0.8 at each epoch. Meanwhile, first 2 methods showed more robust results with technique of learning rate decrease during plateau.

The models were compiled with RMSProp while hyperparameter optimization suggested SGD with momentum as an optimizer [22] for 3D CNN architecture.

Further details regarding the training process and model architectures can be found on corresponding author's GitHub page[1].

## IV. RESULTS

This section will demonstrate the performance of different algorithms on the given dataset. The comparison is made on the basis of the main metrics: accuracy and f1 score. The effect of the size of the training data is investigated at the end of this section.

*A. Comparative analysis*

Overall, the results suggest that the combined approach of MobileNetV2 with BiLSTM layers is more accurate for the task at hand compared to the other models. All key metrics about the models are tabulated in Table 2.

*Table 2. Performance of four architectures on training-fraction dataset*

| Model | Accuracy | F1-Score (Class 0) | F1-Score (Class 1) |
|---|---|---|---|
| 3D CNN | 0.74 | 0.75 | 0.73 |
| 2D CNN & LSTM | 0.74 | 0.73 | 0.76 |
| InceptionV3 & BiLSTM | 0.84 | 0.83 | 0.84 |
| MobileNetV2 & BiLSTM | 0.87 | 0.86 | 0.88 |

[1] https://github.com/DDursun/Violence-Detection

The first and second approaches achieved 74% accuracy, demonstrating limitations in addressing the complexity of the dataset. The classification accuracy of InceptionV3 is 84%. Overall results are satisfactory considering the size and the complexity of the dataset. The last approach utilizes MobileNetV2 as feature extractor. This approach shows better accuracy compared to the InceptionV3 and Bi-LSTM combination. The results are promising however the model contains 40 cases where the nonviolent act is classified as violence. This can trigger pointless alarms and resource allocation. The model also outperformed the InceptionV3 in training time due to its lightweight architecture.

*B. Does more data beat better algorithms?*

Kaisler et al. [23] claim this question to be one of the unsolved mysteries in the advanced analytics space which also inspired the testing of this phenomenon in this problem. To answer this question, videos used for training were increased from 500 to 1600. An average of 6.3% improvement has been achieved while using all the available data. Fig. 4 illustrates the performance of algorithms in training-fraction and training-full datasets.

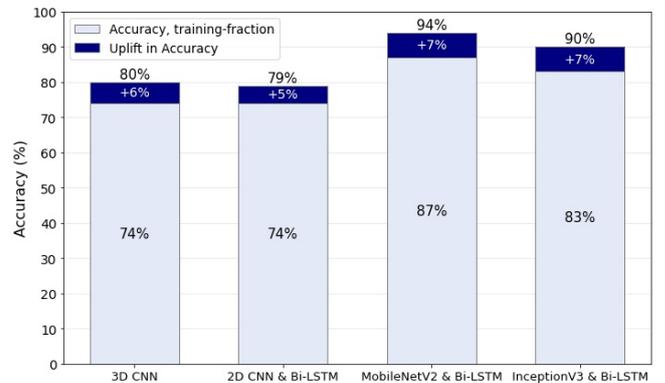

*Fig. 4. Performance of algorithms in training-full dataset*

Although complex architectures with pre-trained weights have more robust performance, the results suggest that increase of the dataset volume in violence detection problems, results in considerable uplift in accuracy.

MobileNetV2 and Bi-LSTM pairing outperformed their counterparts in both sections of experimentation. The confusion matrix is also plotted for the better understanding of classification performance of our best model. As can be seen from Fig 5., the model correctly classified 377 videos out of 400.

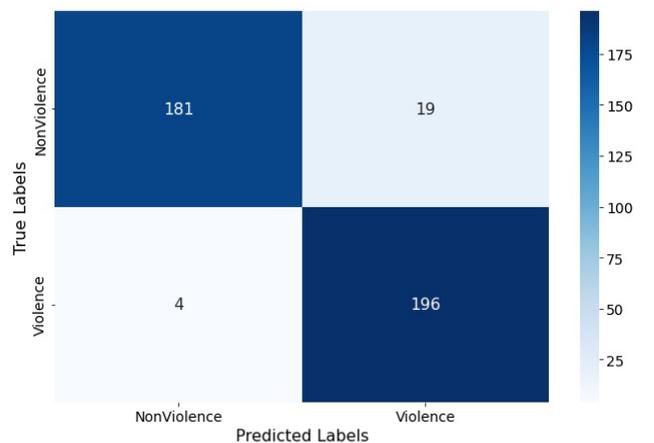

*Fig 5. Performance of MobileNetV2&Bi-LSTM on training-full dataset*

## V. Conclusions and Future work

In this research, four advanced deep learning architectures were compared on a dataset of considerable complexity compared to those utilized in prior studies. The analysis demonstrated that each model achieved a level of accuracy that could be deemed satisfactory in a general context. However, given the critical implications associated with misclassifications in this domain, the performance of CNN&LSTM and 3D CNN architectures was determined to be suboptimal since they failed to sufficiently identify the features within the tested dataset.

The integrated model comprising MobileNetV2 and BiLSTM network demonstrated superior performance over its counterparts in both experiments. The hybrid model emerged as the most robust solution for the classification task in the considered dataset. Moreover, the MobileNetV2 framework offers a computational efficiency that surpasses that of the InceptionV3, endorsing its suitability where computational resources are limited.

In terms of importance of data size, our experimentation proved that considerable improvement can be achieved by increasing the amount of training data. Considering the diversity of videos that models can encounter more data can potentially help to create robust classifiers. There are several available datasets that already have been investigated by researchers. This is one of the potential improvements in this area of computer vision – a collection of diverse, high volume and, high-quality datasets for violence recognition.

Although we experienced an uplift in training performance through the usage of transfer learning methods, the evaluation of proposed approaches can include quantitative analysis of computational demand and time.

Lastly, the recent survey by Kaur and Singh [24] identified low performance of violence detection algorithms on cross datasets. The future work of ours will include testing of the proposed methodology on combined datasets to ensure generalization of model for various situations.

## Acknowledgements

A heartfelt thank you to A. Adamov for his invaluable support throughout the preparation of this paper.